\RequirePackage{fix-cm}
\documentclass[10pt,twocolumn,letterpaper]{article}
\usepackage[pagenumbers]{cvpr}
\usepackage{graphicx}
\usepackage{amsmath,amssymb}
\usepackage{booktabs,array,tabularx,multirow}
\usepackage{xcolor}
\usepackage{tikz}
\usetikzlibrary{arrows.meta,positioning,fit,backgrounds,calc,shapes.geometric}
\usepackage{microtype}
\definecolor{pcBlue}{HTML}{0072B2}\definecolor{pcOrange}{HTML}{E69F00}
\definecolor{pcGreen}{HTML}{009E73}\definecolor{pcPurple}{HTML}{CC79A7}
\definecolor{pcSky}{HTML}{56B4E9}\definecolor{pcVerm}{HTML}{D55E00}
\definecolor{pcGrey}{HTML}{6E6E6E}
\usepackage{caption}
\usepackage{xspace}
\newcommand{\method}{PlaceTally\xspace}

\newcommand{\ind}{\mathbf{1}}
\newcolumntype{Y}{>{\raggedright\arraybackslash}X}

\usepackage{colortbl}
\usepackage{soul}
\definecolor{cvprblue}{rgb}{0.21,0.49,0.74}
\usepackage[pagebackref,breaklinks,colorlinks,allcolors=cvprblue]{hyperref}
\hypersetup{pdftitle={PlaceTally: A Test-Time Memory Model for Constructing Place Graphs},pdfauthor={Alibay Osmanli, Shaogang Gong}}
\sethlcolor{yellow}
\soulregister{\cite}{7}
\soulregister{\ref}{7}
\title{PlaceTally: A Test-Time Memory Model\\for Constructing Place Graphs}
\author{Alibay Osmanli \qquad Shaogang Gong\\
Queen Mary University of London\\
{\tt\small \{a.osmanli, s.gong\}@qmul.ac.uk}\\[0.4em]
{\small Code, data and supplementary material: \url{https://alibayosmanli.github.io/placetally}}}

\begin{document}
\twocolumn[{\maketitle\vspace{-2mm}
\centering
\begin{tikzpicture}[x=1cm,y=1cm,>={Latex[length=1.5mm,width=1.15mm]}]\def\fA{\fontsize{7}{8}\selectfont}\def\fB{\fontsize{6.4}{7.4}\selectfont}\def\fC{\fontsize{5.8}{6.8}\selectfont}\def\fD{\fontsize{5.2}{6}\selectfont}
\node[inner sep=0,anchor=south west] at (2.300,2.75) {\includegraphics[width=0.900cm,height=1.2cm]{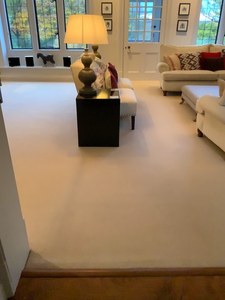}};
\draw[pcBlue,line width=1pt] (2.300,2.75) rectangle (3.200,3.950);
\node[fill=pcBlue,text=white,font=\fC\bfseries,inner sep=.8pt,minimum size=3mm,anchor=north west] at (2.300,3.950) {A};
\node[inner sep=0,anchor=south west] at (3.390,2.75) {\includegraphics[width=1.600cm,height=1.2cm]{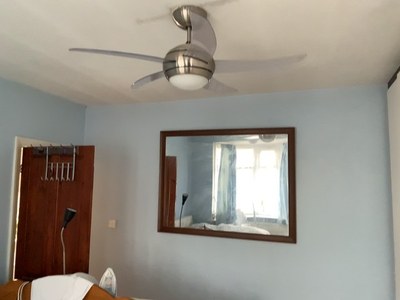}};
\draw[pcOrange,line width=1pt] (3.390,2.75) rectangle (4.990,3.950);
\node[fill=pcOrange,text=white,font=\fC\bfseries,inner sep=.8pt,minimum size=3mm,anchor=north west] at (3.390,3.950) {B};
\node[inner sep=0,anchor=south west] at (5.180,2.75) {\includegraphics[width=0.900cm,height=1.2cm]{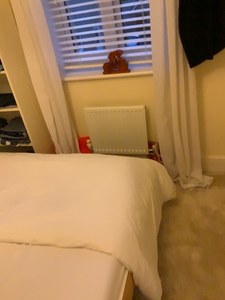}};
\draw[pcGreen,line width=1pt] (5.180,2.75) rectangle (6.080,3.950);
\node[fill=pcGreen,text=white,font=\fC\bfseries,inner sep=.8pt,minimum size=3mm,anchor=north west] at (5.180,3.950) {C};
\node[inner sep=0,anchor=south west] at (6.270,2.75) {\includegraphics[width=1.600cm,height=1.2cm]{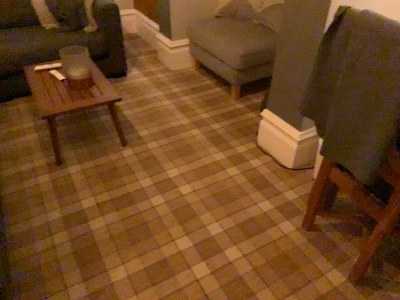}};
\draw[pcPurple,line width=1pt] (6.270,2.75) rectangle (7.870,3.950);
\node[fill=pcPurple,text=white,font=\fC\bfseries,inner sep=.8pt,minimum size=3mm,anchor=north west] at (6.270,3.950) {D};
\node[inner sep=0,anchor=south west] at (8.060,2.75) {\includegraphics[width=1.600cm,height=1.2cm]{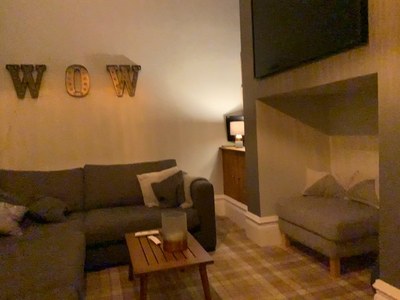}};
\draw[pcPurple,line width=1pt] (8.060,2.75) rectangle (9.660,3.950);
\node[fill=pcPurple,text=white,font=\fC\bfseries,inner sep=.8pt,minimum size=3mm,anchor=north west] at (8.060,3.950) {D};
\node[inner sep=0,anchor=south west] at (9.850,2.75) {\includegraphics[width=0.900cm,height=1.2cm]{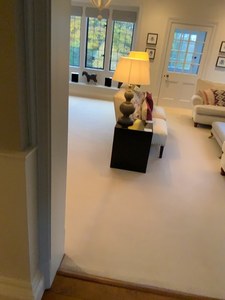}};
\draw[pcBlue,line width=1pt] (9.850,2.75) rectangle (10.750,3.950);
\node[fill=pcBlue,text=white,font=\fC\bfseries,inner sep=.8pt,minimum size=3mm,anchor=north west] at (9.850,3.950) {A};
\node[anchor=west,font=\fB\itshape,inner sep=0] at (0,3.55) {``How many tables?"};
\node[anchor=west,font=\fC,text=black!65,inner sep=0] at (0,3.25) {one stream of 13 minutes};
\fill[pcBlue] (2.312,2.000) rectangle (5.186,2.400);
\node[text=white,font=\fA\bfseries,inner sep=0] at (3.749,2.200) {3};
\node[font=\fD\bfseries,text=pcBlue!85!black,inner sep=0] at (3.749,2.530) {A};
\fill[pcOrange] (5.210,2.000) rectangle (6.060,2.400);
\node[text=white,font=\fA\bfseries,inner sep=0] at (5.635,2.200) {0};
\node[font=\fD\bfseries,text=pcOrange!85!black,inner sep=0] at (5.635,2.530) {B};
\fill[pcGreen] (6.084,2.000) rectangle (6.745,2.400);
\node[text=white,font=\fA\bfseries,inner sep=0] at (6.414,2.200) {0};
\node[font=\fD\bfseries,text=pcGreen!85!black,inner sep=0] at (6.414,2.530) {C};
\fill[pcPurple] (6.769,2.000) rectangle (7.884,2.400);
\node[text=white,font=\fA\bfseries,inner sep=0] at (7.327,2.200) {3};
\node[font=\fD\bfseries,text=pcPurple!85!black,inner sep=0] at (7.327,2.530) {D};
\fill[pcPurple] (7.908,2.000) rectangle (8.681,2.400);
\node[text=white,font=\fA\bfseries,inner sep=0] at (8.295,2.200) {2};
\node[font=\fD\bfseries,text=pcPurple!85!black,inner sep=0] at (8.295,2.530) {D};
\fill[pcBlue] (8.705,2.000) rectangle (10.738,2.400);
\node[text=white,font=\fA\bfseries,inner sep=0] at (9.721,2.200) {3};
\node[font=\fD\bfseries,text=pcBlue!85!black,inner sep=0] at (9.721,2.530) {A};
\fill[black!15] (2.312,1.480) rectangle (10.738,1.880);
\node[text=black,font=\fA\bfseries,inner sep=0] at (6.525,1.680) {4};
\fill[pcBlue] (2.312,0.960) rectangle (5.186,1.360);
\node[text=white,font=\fA\bfseries,inner sep=0] at (3.749,1.160) {3};
\fill[pcOrange] (5.210,0.960) rectangle (6.060,1.360);
\node[text=white,font=\fA\bfseries,inner sep=0] at (5.635,1.160) {0};
\fill[pcGreen] (6.084,0.960) rectangle (6.745,1.360);
\node[text=white,font=\fA\bfseries,inner sep=0] at (6.414,1.160) {0};
\fill[pcPurple] (6.769,0.960) rectangle (8.681,1.360);
\node[text=white,font=\fA\bfseries,inner sep=0] at (7.725,1.160) {2};
\fill[pcBlue] (8.705,0.960) rectangle (10.738,1.360);
\node[anchor=east,font=\fA,inner sep=0] at (2.200,2.200) {Sum over visits};
\node[anchor=east,font=\fA,inner sep=0] at (2.200,1.680) {Whole-video reading};
\node[anchor=east,text=pcBlue,font=\fA\bfseries,inner sep=0] at (2.200,1.160) {PlaceTally};
\draw[->,pcBlue,line width=.8pt,rounded corners=1.2mm] (9.721,0.940) -- (9.721,.5) -- (3.749,.5) -- (3.749,0.940);
\node[fill=white,font=\fB,inner sep=1.5pt] at (6.735,.5) {return joined to A: counted once};
\node[anchor=east,font=\fontsize{8.5}{9}\selectfont,text=black!70,inner sep=0] at (11.350,2.200) {11};
\draw[pcVerm,line width=1.1pt,line cap=round] (11.550,2.120) -- (11.710,2.280) (11.550,2.280) -- (11.710,2.120);
\node[anchor=east,font=\fontsize{8.5}{9}\selectfont,text=black!70,inner sep=0] at (11.350,1.680) {4};
\draw[pcVerm,line width=1.1pt,line cap=round] (11.550,1.600) -- (11.710,1.760) (11.550,1.760) -- (11.710,1.600);
\node[anchor=east,font=\fontsize{8.5}{9}\selectfont\bfseries,text=pcBlue,inner sep=0] at (11.350,1.160) {5};
\draw[pcGreen!80!black,line width=1.1pt,line cap=round,line join=round] (11.540,1.160) -- (11.610,1.080) -- (11.740,1.260);
\node[anchor=east,font=\fB,inner sep=0] at (11.750,.5) {truth 5};
\draw[black!70,line width=.4pt] (13.15,3.92) -- (13.15,0.62) -- (17.25,0.62);
\draw[black!70,line width=.4pt] (13.305,0.62) -- (13.305,0.57) node[below=-.5pt,font=\fC,inner sep=1pt] {1};
\draw[black!70,line width=.4pt] (13.15,0.738) -- (13.1,0.738) node[left=-.5pt,font=\fC,inner sep=1pt] {1};
\draw[black!70,line width=.4pt] (14.328,0.62) -- (14.328,0.57) node[below=-.5pt,font=\fC,inner sep=1pt] {2};
\draw[black!70,line width=.4pt] (13.15,1.517) -- (13.1,1.517) node[left=-.5pt,font=\fC,inner sep=1pt] {2};
\draw[black!70,line width=.4pt] (15.350,0.62) -- (15.350,0.57) node[below=-.5pt,font=\fC,inner sep=1pt] {4};
\draw[black!70,line width=.4pt] (13.15,2.295) -- (13.1,2.295) node[left=-.5pt,font=\fC,inner sep=1pt] {4};
\draw[black!70,line width=.4pt] (16.373,0.62) -- (16.373,0.57) node[below=-.5pt,font=\fC,inner sep=1pt] {8};
\draw[black!70,line width=.4pt] (13.15,3.074) -- (13.1,3.074) node[left=-.5pt,font=\fC,inner sep=1pt] {8};
\node[font=\fB,inner sep=0] at (15.200,0.26) {True count};
\node[font=\fB,inner sep=0,rotate=90] at (12.73,2.270) {Mean answer};
\draw[black,line width=.5pt,dash pattern=on 2pt off 1.5pt] (13.230,0.681) -- (17.219,3.718);
\fill[pcVerm,opacity=.16] (13.305,1.746) -- (14.328,2.168) -- (15.112,2.539) -- (15.956,3.083) -- (17.009,3.779) -- (17.009,3.562) -- (15.956,2.913) -- (15.112,2.400) -- (14.328,2.028) -- (13.305,1.463) -- cycle;
\draw[pcVerm,line width=.9pt] (13.305,1.602) -- (14.328,2.093) -- (15.112,2.472) -- (15.956,3.001) -- (17.009,3.676);
\node[rectangle,fill=pcVerm,inner sep=0,minimum size=1.5mm] at (13.305,1.602) {};
\node[rectangle,fill=pcVerm,inner sep=0,minimum size=1.5mm] at (14.328,2.093) {};
\node[rectangle,fill=pcVerm,inner sep=0,minimum size=1.5mm] at (15.112,2.472) {};
\node[rectangle,fill=pcVerm,inner sep=0,minimum size=1.5mm] at (15.956,3.001) {};
\node[rectangle,fill=pcVerm,inner sep=0,minimum size=1.5mm] at (17.009,3.676) {};
\fill[pcBlue,opacity=.16] (13.305,1.078) -- (14.328,1.638) -- (15.112,2.122) -- (15.956,2.654) -- (17.009,3.370) -- (17.009,3.167) -- (15.956,2.469) -- (15.112,1.963) -- (14.328,1.503) -- (13.305,0.717) -- cycle;
\draw[pcBlue,line width=.9pt] (13.305,0.902) -- (14.328,1.568) -- (15.112,2.045) -- (15.956,2.563) -- (17.009,3.276);
\node[circle,fill=pcBlue,inner sep=0,minimum size=1.5mm] at (13.305,0.902) {};
\node[circle,fill=pcBlue,inner sep=0,minimum size=1.5mm] at (14.328,1.568) {};
\node[circle,fill=pcBlue,inner sep=0,minimum size=1.5mm] at (15.112,2.045) {};
\node[circle,fill=pcBlue,inner sep=0,minimum size=1.5mm] at (15.956,2.563) {};
\node[circle,fill=pcBlue,inner sep=0,minimum size=1.5mm] at (17.009,3.276) {};
\fill[pcGrey,opacity=.16] (13.305,1.256) -- (14.328,1.599) -- (15.112,1.811) -- (15.956,2.129) -- (17.009,2.761) -- (17.009,2.517) -- (15.956,1.943) -- (15.112,1.710) -- (14.328,1.522) -- (13.305,1.051) -- cycle;
\draw[pcGrey,line width=.9pt] (13.305,1.156) -- (14.328,1.559) -- (15.112,1.762) -- (15.956,2.029) -- (17.009,2.637);
\node[diamond,fill=pcGrey,inner sep=0,minimum size=1.5mm] at (13.305,1.156) {};
\node[diamond,fill=pcGrey,inner sep=0,minimum size=1.5mm] at (14.328,1.559) {};
\node[diamond,fill=pcGrey,inner sep=0,minimum size=1.5mm] at (15.112,1.762) {};
\node[diamond,fill=pcGrey,inner sep=0,minimum size=1.5mm] at (15.956,2.029) {};
\node[diamond,fill=pcGrey,inner sep=0,minimum size=1.5mm] at (17.009,2.637) {};
\draw[pcVerm,line width=.9pt] (13.27,3.900) -- (13.57,3.900);
\node[anchor=west,font=\fB,inner sep=0,text=pcVerm!90!black] at (13.63,3.900) {Sum over visits};
\draw[pcBlue,line width=.9pt] (13.27,3.600) -- (13.57,3.600);
\node[anchor=west,font=\fB\bfseries,inner sep=0,text=pcBlue!90!black] at (13.63,3.600) {PlaceTally};
\draw[pcGrey,line width=.9pt] (13.27,3.300) -- (13.57,3.300);
\node[anchor=west,font=\fB,inner sep=0,text=pcGrey!90!black] at (13.63,3.300) {Whole-video reading};
\draw[black,line width=.5pt,dash pattern=on 2pt off 1.5pt] (13.27,3.000) -- (13.57,3.000);\node[anchor=west,font=\fB,inner sep=0] at (13.63,3.000) {true count};
\end{tikzpicture}

\captionsetup{hypcap=false}

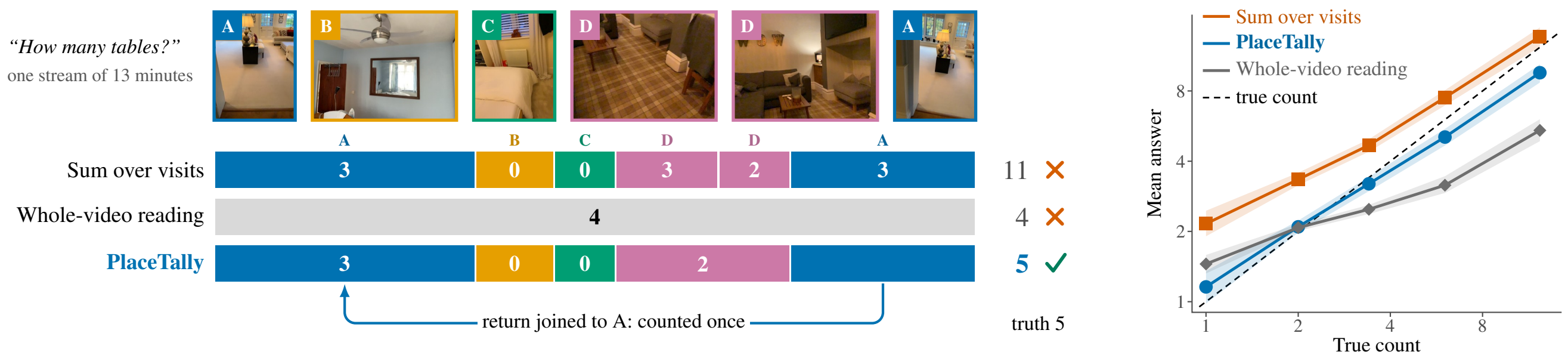
\captionof{figure}{Why do video models miscount when the camera returns? Reading the whole video gives a few small answers, adding up visits counts each return again, and \method counts each place once. Left: one ARKit-Revisit stream of 13 minutes with six recordings of four rooms, where room D is recorded twice in a row and room A returns at the end; numbers are the same model's counts. Right: mean answer against binned true count on all 460 test questions, with 95\% bands. Every question here has an object seen on a return (Sec.~\ref{sec:protocol}).
\label{fig:teaser}}
\vspace{3mm}}]

\begin{abstract}
Counting distinct objects in a video is hard when the camera returns to places it has seen. A vision-language model that reads the whole video answers with a few small numbers, and one that reads it in parts counts an object again in every part that shows it. Object trackers need the category before the video is read, or a fixed vocabulary.
We observe that a memory for counting does not need to know what it is looking for, only where it has been: objects rarely leave a room between visits, so parts of a video grouped by place share almost no objects, and a place can be recognised before any question is asked.
Building on this, we introduce \method, a question-blind test-time memory model that constructs a place graph for each video.
Each video fragment joins the place of its best earlier match or opens a new place, because a strong match is rarely wrong while a true return often matches weakly.
A frozen vision-language model counts each place once from at most 128 of its frames, so every read has a bounded input and the memory grows with the number of places; the graph is built 15 times faster than real time.
On RevisitCount, our benchmark of real room re-recordings, it scores 62.5\% mean relative accuracy on test rooms and 61.2\% on held-out rooms, against 51.2\% and 44.3\% for the same model reading the whole video.
It stays level as rooms return more or less often, while a fixed correction of summed visits moves from 38.5\% to 61.4\%.
\end{abstract}
\section{Introduction}
\label{sec:intro}
Object counting~\cite{ranjan2021famnet,amininaieni2024countgd} estimates how many instances of a category appear in an image; in video, it must also decide which observations show the same object~\cite{han2022drvic,amini2025countvid}.
This decision is hardest when the camera returns to a place it has already seen, as an assistant or a robot does~\cite{yang2024thinking,lin2025ostbench}, because every return shows the same objects again.
The question also comes later, so the memory has to be constructed before it knows what will be counted.

Existing methods fail for two opposing reasons (Fig.~\ref{fig:teaser}): reading the whole video at once under-counts, and adding up parts over-counts.
A vision-language model asked about a whole video answers with a few small numbers, a weakness already reported on images~\cite{qharabagh2026lvlmcount} and on synthetic videos~\cite{chapus2026countvideobench}: on our benchmark, 87\% of its answers are 1, 2 or 3, and for nine or more objects it answers 5.4 on average.
Reading in parts keeps each count small, but adding up counts an object again in every part that shows it: the counting mode of Cambrian-S~\cite{yang2025cambrian} works by adding up over surprise-segmented pieces, which fails when a video repeats~\cite{udandarao2025supersensing}, and on our videos predicts a median of 8.6 times the true count.
Methods that track objects need the objects first. CountVid~\cite{amini2025countvid} needs the category known before the video is processed and tracks it for every question, StaMina~\cite{liu2026stamina} registers its targets when the stream starts, and VideoAgent~\cite{fan2024videoagent} and DirectMe~\cite{wang2026directme} keep objects from a detector vocabulary.

The core problem is that a model can count only a few objects at a time; a long video must be counted in parts, and parts can be added up only if they share no objects.
How much they share depends on which rooms this camera came back to and must be worked out for each video. Finding parts that share no objects appears to need the objects themselves: on images, LVLM-Count~\cite{qharabagh2026lvlmcount} first segments the target so that no object is cut.
We observe that in video it does not. Objects rarely leave a room between visits, so parts grouped by place share almost no objects, and ``have I seen this object?'' becomes ``have I been here before?''.
This question names no category.
Answering it is a computational problem of its own: the video must be partitioned online into an unknown number of places, with no depth or camera path, when two views of one room can share no visible surface and different rooms look alike. Each decision is final: a split counts a room's objects twice, and a wrong merge asks the model to count two rooms as one.

To overcome these problems, our idea is to introduce a test-time memory model that is built for each test video instance: a place graph, constructed before any question is asked.
As the category is discovered only afterwards, the same memory must serve any category (open-vocabulary); we call this question-blind memory. It does not need to know what it is looking for, only where it has been.
It is built per video and not accumulated over a dataset, because only this video's returns decide what can be added up, and rooms in other videos add look-alikes; it is cheap enough to rebuild for every video.
Our main contributions are:
\begin{itemize}\setlength{\itemsep}{1pt}\setlength{\parskip}{0pt}\setlength{\topsep}{2pt}
\item We show that counting across returns can be posed as partitioning a video into places. Earlier work groups video by place to map a kitchen or a building~\cite{nagarajan2020egotopo,hughes2022hydra}, or sums counts over image parts found by segmenting the target~\cite{qharabagh2026lvlmcount}; with the place as the unit of counting identity, no category, vocabulary or per-question tracking is needed. With the same frozen model, this lifts mean relative accuracy (MRA~\cite{yang2024thinking}) from 51.2\% to 62.5\% on real returns, within 2.6\% of the score with true visits, from 44.3\% to 61.2\% on held-out rooms evaluated once, and from 10.5\% to 56.3\% at nine or more objects, where whole-video reading is weakest.
\item We introduce a test-time memory model of where the camera has been: a place graph constructed per video before any question. It must recognise a place online with no category, depth or camera path, although one room can look different and two rooms alike. We find the appearance evidence one-sided (a strong match is rarely wrong, a true return often matches weakly), so one decision per fragment on its strongest link is enough, about nine on an 11-minute video, while summing all links over-splits places. With at most $K$ frames per place, $K$ sets the memory, each read's input and the answer time, and the graph is built 15 times faster than real time (Sec.~\ref{sec:size}).
\item We introduce RevisitCount. The eight benchmarks we inspected have videos of up to two hours, up to 10,000 questions and, in five cases, a blind check (Table~\ref{tab:benchmark}), but none reports a return to a room as a separate recording, so adding up parts is not penalised and a model can score without recognising a place: at 60 minutes, a prior from the video's length reaches 80.8\% on VSI-SUPER-Count but 17.5\% on ours. RevisitCount records every return separately, counts each object once and reports blind baselines on every split; one split of 248 rooms was held out and evaluated once.
\end{itemize}
\section{Related Work}
\label{sec:related}
\begin{table}[t]
\centering\footnotesize
\setlength{\tabcolsep}{1.0pt}
\begin{tabular}{@{}lllll@{}}
\toprule
Method & Identity of & \shortstack[l]{Before\\question} & \shortstack[l]{Named\\in advance} & \shortstack[l]{Depth,\\path} \\
\midrule
CountVid~\cite{amini2025countvid} & target objects & no & the target & no \\
StaMina~\cite{liu2026stamina} & objects, events & yes & targets & no \\
VideoAgent~\cite{fan2024videoagent} & objects & yes & detector & no \\
Emb.\ VideoAgent~\cite{fan2025embodiedvideoagent} & 3D objects & yes & detector & yes \\
DirectMe~\cite{wang2026directme} & objects, places & yes & Objects365~\cite{shao2019objects365} & est. \\
Hydra~\cite{hughes2022hydra} & objects, rooms & yes & labels & yes \\
Ego-Topo~\cite{nagarajan2020egotopo} & zones & yes & none & no \\
\method & places & yes & none & no \\
\bottomrule
\end{tabular}
\caption{How methods decide identity. Depth, path: depth or camera path used (est.: estimated). Cells come from the primary papers. None of them uses place identity to stop double counting.}
\label{tab:identity}
\end{table}

\paragraph{Counting with vision-language models.}
Class-agnostic counters use exemplars, detection or open-world prompts~\cite{ranjan2021famnet,liu2022countr,djukic2023loca,pelhan2024dave,pelhan2024geco,amininaieni2024countgd,amininaieni2026countgdpp}.
Vision-language models count poorly without grounding~\cite{paiss2023count,deng2026holocount,elshangiti2026misaligned}: they concentrate their answers on preferred integers~\cite{chapus2026countvideobench} and weaken as counts grow. Pointing at objects first helps~\cite{deitke2024molmo,alghisi2026point,clark2026molmo2}, though not inside our places (Supp.~D.9).
In video, DR.VIC~\cite{han2022drvic} and CountVid~\cite{amini2025countvid} match objects across frames, StaMina~\cite{liu2026stamina} learns when a count should change, and benchmarks report failures on long videos~\cite{tsuchiya2026ecbench} and on objects that reappear~\cite{chen2026tocbench}.
LVLM-Count~\cite{qharabagh2026lvlmcount} divides an image around the segmented target and sums the parts. We share its logic but find parts differently: places need no category or segmentation, so a question-blind memory can come first.

\paragraph{Memory for long and streaming video.}
Token memories compress what the model has seen~\cite{song2024moviechat,he2024malmm,zhang2025flashvstream,yang2025streammem} or retrieve cached keys when a question arrives~\cite{di2025rekv}; both accumulate without deciding which parts show the same place, so a count over them must still find every duplicate. Agent memories keep re-identified tracks~\cite{fan2024videoagent}, persistent 3D objects with depth and pose~\cite{fan2025embodiedvideoagent}, image snapshots~\cite{yang2025threedmem} or captions indexed by position and time~\cite{anwar2024remembr}. Spatial models add 3D encoders~\cite{wu2025spatialmllm,fan2025vlm3r} or keep 3D-aware maps and memories~\cite{zheng2026spatialmem,gu2026spacemind,liu2026vlmem}, and Spatial-TTT~\cite{liu2026spatialttt} stores evidence in weights updated at test time.
The closest counting memory is DirectMe~\cite{wang2026directme}, which builds object and place nodes before the question and reports object-quantity accuracy (Table~\ref{tab:identity}); its counting identity still comes from object tracks, with estimated depth and pose and a fixed vocabulary, and on our streams its map holds far more entries than objects (Sec.~\ref{sec:others}).
In the place graph of \method, a place is the unit of identity, and the memory holds no objects at all.

\paragraph{Place recognition and re-identification.}
Visual place recognition retrieves views of one place~\cite{arandjelovic2016netvlad,keetha2023anyloc,berton2025megaloc}, and loop closure recognises a return to a mapped place~\cite{galvez2012dbow}. FAB-MAP~\cite{cummins2008fabmap} adds an explicit new-place option. One place can look different while different places look alike~\cite{lowry2016survey}. Robotics work grouped images into rooms by graph clustering~\cite{zivkovic2007rooms} and segmented image streams into places by change-point detection~\cite{ranganathan2012pliss}, and Ego-Topo~\cite{nagarajan2020egotopo} turns egocentric video into visits to the zones of a kitchen to learn what each zone affords. Topological maps~\cite{savinov2018sptm,chaplot2020nts,suomela2024placenav,wang2026cross} and scene graphs~\cite{hughes2022hydra,gu2024conceptgraphs,werby2024hovsg,zhang2024msg} link views, objects, places or rooms for navigation. Person re-identification matches people across cameras and times~\cite{gong2014reid,ye2022reid}, and sports analytics re-identifies players and locates them on the pitch~\cite{somers2024gsr}.
For counting, re-identifying a returning room replaces object identity. It is harder than re-identifying a person, as two views of one room can share no visible surface, and unlike the zones of a pitch, places must be found in each video.

\paragraph{Graph partitioning.}
Tracking casts association as graph partitioning~\cite{braso2020solver,cetintas2023sushi}, and correlation clustering~\cite{bansal2004correlation} gives the signed objective; its online form merges clusters on summed evidence~\cite{mathieu2010online}. Coreference resolution links each mention to one best antecedent~\cite{lee2017coref}. GraphThinker~\cite{cheng2026graphthinker} builds an event graph of events and entities for grounded video reasoning; our nodes are places, built before any question.
We keep the graph and replace the sum with one best link per fragment.
\section{Methodology}
\label{sec:method}
\begin{figure*}[t]
\centering
\input{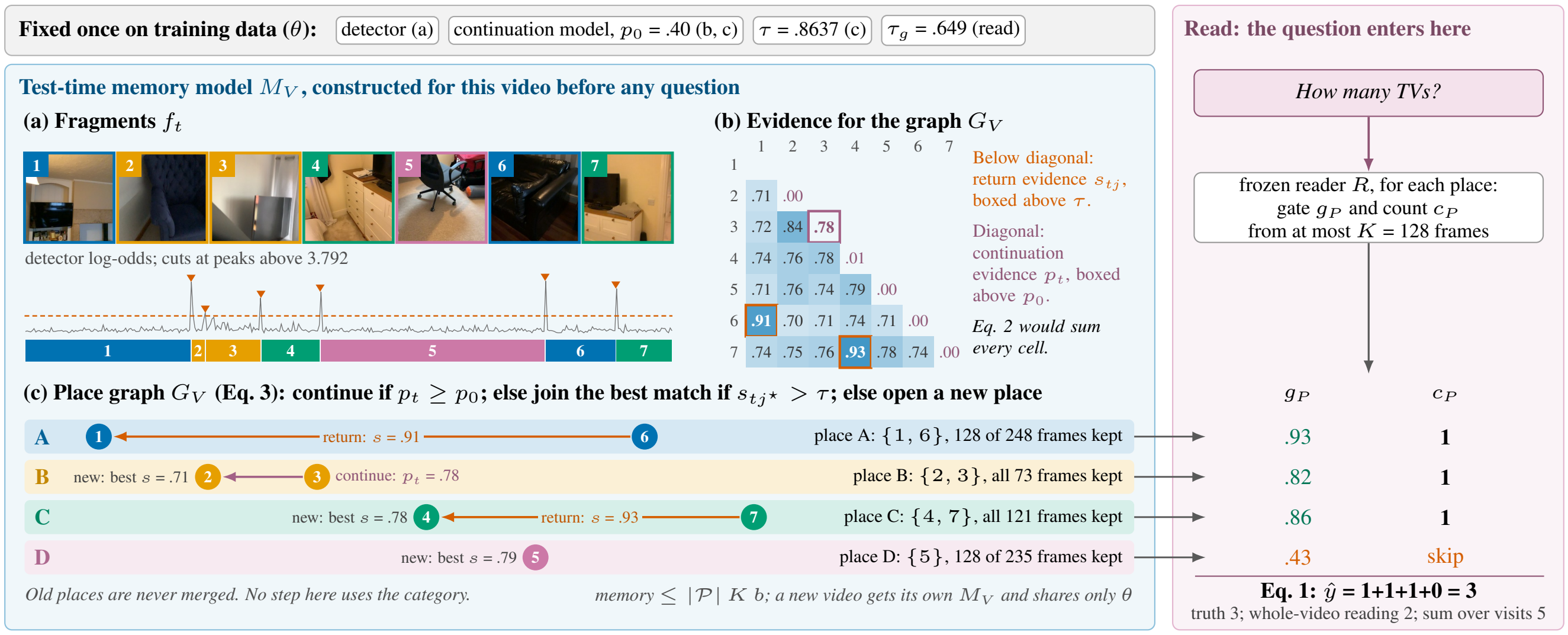}
\caption{A \method memory is constructed per video instance before the question and read after it, shown on one real ARKit-Revisit test stream. (a) The detector cuts the stream into fragments. (b) Continuation and return evidence between fragments, which Eq.~\ref{eq:partition} would sum. (c) The place graph $G_V$: Eq.~\ref{eq:rule} takes one decision per fragment (node), the kept links are its edges, and each lane is a place. Read: the category enters only here; each place is gated and counted, and the counts are summed once per place (Eq.~\ref{eq:read}).
}
\label{fig:method}
\end{figure*}
In this section, we define the task and then describe how \method constructs and reads a question-blind place graph for each video at test time (Fig.~\ref{fig:method}).

\paragraph{Counting as a sum over parts.}
A stream $V$ arrives frame by frame, and a category $q$ is asked afterwards. The target is the number of distinct annotated instances of $q$ in $V$. Whole-video reading computes $\hat y=R(V,q)$ with a reader $R$, whose answer stays small as the truth grows (Sec.~\ref{sec:oneread}). Reading in parts computes $\hat y=\sum_s R(s,q)$, which is right only if no object appears in two parts; we call such a partition object-disjoint. Object memories compute $|\{o\in\mathcal O:\operatorname{cat}(o)=q\}|$, which needs the object set $\mathcal O$.
Our working assumption is that objects stay in one place between visits. Parts grouped by place are then object-disjoint for every $q$, and the task reduces to finding the places.
\method therefore constructs, for each video, a memory model $M_V=W_\theta(V)$, a place graph $G_V$ with the frames of each place, where $W_\theta$ is the construction below with parameters $\theta$. It answers with
\begin{equation}
\hat y(q)=\sum_{P\in\mathcal P}\ind\!\left[g_P\ge\tau_g\right]c_P ,
\label{eq:read}
\end{equation}
where $\mathcal P$ are the places of $G_V$, $c_P$ is the reader's count from at most $K$ frames of $P$, and $g_P$, the presence gate, is the probability that the reader answers yes to ``is $q$ visible?'', which keeps a place without the object out of the sum. $\theta$ is fixed on training data, which share no recording with any test stream; $G_V$ is inferred per test video.

\paragraph{One memory per video.}
What the parts of a video share depends on which rooms this camera came back to; no number fixed across videos can stand in for it, and a rate estimated per video needs its places anyway (Sec.~\ref{sec:correction}). Rooms in other videos only add look-alikes (Sec.~\ref{sec:oneread}), so nothing carries over between videos except $\theta$. The design handles two risks: look-alike rooms inside a video, with a strict threshold and no merging of old places, and memory that grows with the video, by keeping at most $K$ frames per place. A third remains: a $\theta$ fitted on one camera may not hold on another (Sec.~\ref{sec:behind}).

\paragraph{Fragments.}
The nodes of $G_V$ are fragments $f_1,\dots,f_n$, cut from the stream at 5 fps. A missed cut joins two rooms for good, while a false cut, inside one recording, can be repaired, so the detector cuts generously. It is a logistic regression on six similarities between neighbouring frames, among them SigLIP~2~\cite{tschannen2025siglip2} (Supp.~B), fitted on streams joined from other ARKitScenes~\cite{baruch2021arkit} recordings, and it cuts at peaks where its log-odds exceed 3.792.

\paragraph{Continuation evidence.}
Generous cuts over-split a recording: 496 of 1,791 cuts on the training streams are false.
Appearance cannot repair this, because two fragments of one recording often look at different walls: on the training streams, 93\% of fragment pairs from one recording score below the return threshold defined next.
Time can. At each cut, a continuation model, a logistic regression on six statistics fitted on training streams, gives $p_t$, the probability that the cut before $f_t$ is false. One statistic (Supp.~B) is a learned similarity: SigLIP~2 features under a linear map, trained on other recordings to match 8-s windows of one recording.

\paragraph{Return evidence.}
Time cannot connect a return, which is a new recording, so appearance has to. The return evidence $s_{tj}$ is the SigLIP~2 chamfer similarity between $f_t$ and an earlier fragment $f_j$ (each frame's best cosine match, averaged both ways), and it is not learned. It is one-sided: a high score is reliable and a low one is not, as a room looks different from another side; on the training streams 40 to 50\% of true-return pairs score below the threshold $\tau$, while 95 to 97\% of pairs above it show the same place (Sec.~\ref{sec:bestlink}).

\paragraph{The best-link rule.}
The standard formulation of identity across frames partitions a graph~\cite{braso2020solver,cetintas2023sushi}. With weights $w_{ij}$ that calibrate the two kinds of evidence as log-odds of the same place, the places would be
\begin{equation}
\mathcal P^\star=\arg\max_{\mathcal P}\sum_{P\in\mathcal P}\ \sum_{i<j;\ f_i,f_j\in P} w_{ij},
\label{eq:partition}
\end{equation}
the objective of correlation clustering~\cite{bansal2004correlation}. This sum counts every low score against the same place, although many true returns score low, and it needs the whole graph, which a stream does not have.
\method keeps the graph and scores each fragment by its strongest link instead of the sum of all its links. Each fragment continues the current place, joins the place of its strongest earlier match, or opens a new place:
\begin{equation}
\ell(f_t)=\begin{cases}\ell(f_{t-1}) & \text{if } p_t\ge p_0,\\ \ell(f_{j^\star}) & \text{else if } s_{tj^\star}>\tau,\\ \text{a new place} & \text{otherwise,}\end{cases}
\label{eq:rule}
\end{equation}
where $j^\star=\arg\max_{j<t}s_{tj}$, $p_0=0.40$ and $\tau=0.8637$.
The care lies in four choices: a strict threshold, one strongest link, continuation first, and no merging of old places.
Eq.~\ref{eq:rule} is exact for an objective of its own: once continuation is decided, each other fragment scores its best link into its own place, $s_{tj}-\tau$, or zero if it opens one, and the rule reaches this maximum for every fragment, online.
It follows antecedent ranking in coreference~\cite{lee2017coref} with FAB-MAP's new-place option~\cite{cummins2008fabmap}. The closest design, Ego-Topo's online zone graph~\cite{nagarajan2020egotopo}, scores a zone by its mean match; we use the strongest link, put continuation first, and count by place.

\paragraph{Size under control.}
Each call to the reader sees at most $K=128$ frames of one place, chosen uniformly at 1 fps, about 13 thousand tokens, whatever the video's length.
Once the graph is built, answering needs only those frames, at most $|\mathcal P|\,K\,b$ for $b$ bytes per frame, so $K$ sets the memory, the input and, through it, the answer time (Sec.~\ref{sec:size}).

\paragraph{What is learned.}
The memory is question-blind because no part of Eq.~\ref{eq:rule} uses $q$. It is test-time because $G_V$ is inferred for each video, and no weights are updated on the test video. What is trained is small: two logistic regressions with six features each and one linear projection, about 295,000 weights, nearly all in the projection, fitted once with labels from recording and room IDs.
The reader is frozen (Qwen3.5-9B~\cite{qwen2026qwen35}; 448-pixel frames, greedy decoding, prompts in Supp.~B), and $\tau_g=0.649$, like $\tau$, was fixed before any test stream existed.
\section{RevisitCount Benchmark}
\label{sec:protocol}
\begin{table}[t]
\centering\footnotesize
\setlength{\tabcolsep}{1.7pt}
\begin{tabular}{@{}lrrlll@{}}
\toprule
Benchmark & Videos & QA & Length & Returns & Blind \\
\midrule
VSI-Bench~\cite{yang2024thinking} & 288 & $>$5,000 & n/r & n/r & yes \\
VSI-SUPER-Count~\cite{yang2025cambrian} & n/r & 200 & 10--120 min & n/r & n/r \\
ReVSI~\cite{zhang2026revsi} & 381 & n/r & n/r & n/r & partial \\
OST-Bench~\cite{lin2025ostbench} & 1,400 & 10,000 & n/r & n/r & chance \\
EC-Bench~\cite{tsuchiya2026ecbench} & 152 & 1,699 & 57 min & n/r & n/r \\
TOC-Bench~\cite{chen2026tocbench} & 1,951 & 2,323 & 29 s & n/r & filter \\
LongSpace~\cite{lang2026longspace} & 445 & 4,073 & 21 min & n/r & n/r \\
SVCBench~\cite{liu2026svcbench} & 406 & 1,000 & to 20+ min & rotation & text-only \\
\midrule
ARKit-Revisit test & 103 & 460 & 11 min & recorded & yes \\
ARKit-Revisit held-out & 87 & 403 & 12 min & recorded & yes \\
3RScan-Revisit test & 134 & 362 & n/a & recorded & yes \\
Long streams & 38, 18 & 202, 68 & 30, 61 min & recorded & yes \\
MultiScan-Revisit & 33 & -- & n/a & recorded & -- \\
\bottomrule
\end{tabular}
\caption{Benchmarks we inspected and RevisitCount (below the line). Returns: how a room is seen again. Blind: blind baseline reported (partial: object-removed videos; filter: only to filter questions). n/r: not reported. Our videos: streams with questions (MultiScan: all streams).}
\label{tab:benchmark}
\end{table}

We further introduce RevisitCount, a benchmark for counting objects across real returns, which we will release with our code. The benchmarks we inspected report no separately recorded return to a room (Table~\ref{tab:benchmark}), so a method can score on them without recognising a place: on VSI-SUPER-Count~\cite{yang2025cambrian}, a baseline that sees only the video's length and category rates fitted on its other durations reaches 80.8\% at 60 minutes, against 32.8\% for \method, while a length prior scores 17.5\% on our 60-minute streams.
Rendered walks do not fill this gap: on HM3D scans~\cite{ramakrishnan2021hm3d} and synthetic HSSD homes~\cite{khanna2024hssd}, a per-category prior fitted on other houses scores 58.4\% and 67.1\%, against 41.6\% for the best blind baseline on ARKit-Revisit test (Supp.~D.4, D.8).

\paragraph{ARKit-Revisit.}
Each stream joins real ARKitScenes recordings~\cite{baruch2021arkit}, one per visit; a return, a later visit to the same room, is a separate recording made the same day, and look-alike distractor rooms are included. We choose the order of rooms; each stream is a constructed walk. Truth comes from annotated 3D boxes: an object counts if it is visible in at least two frames of one visit. Objects are matched across recordings by aligning the scans (Supp.~A).
We keep a category only if at least one of its objects is seen on more than one visit (467 of 828 candidate questions on the test split): every question has an object seen on a return. The 14 categories are furniture, fixtures and appliances.
Each stream has four rooms and two returns or six rooms and three, so about a third of visits are returns; we call this share the return rate, and Sec.~\ref{sec:correction} varies it.
193 training streams, from rooms in no test split, fit the continuation model and the blind baselines. The held-out split, built from 248 rooms used nowhere else, was evaluated once, after the method was fixed. A request error lost seven of the 467 test questions before scoring (Supp.~A).

\paragraph{3RScan-Revisit.}
This split uses rescans of rooms taken months apart, as irregular Tango keyframes~\cite{wald2019rio}, with 60 training and 418 test rooms. Truth is a rendered labelled mesh with the same visibility test. It tests our assumption: 41\% of target objects in both scans were moved, yet only 1.7\% of the 4,787 first-scan objects are listed as removed.

\paragraph{MultiScan-Revisit and long streams.}
MultiScan-Revisit uses iPad video~\cite{mao2022multiscan}, was used for no design choice, and has too few questions for counting, so we evaluate only its places. Long streams, which join ARKit test rooms into 30- and 60-minute streams, are not an independent test.

\paragraph{Metric.}
Strict mean relative accuracy (MRA)~\cite{yang2024thinking} for truth $g$ and answer $p$ is $\frac{1}{10}\sum_{a}\ind[\,|p-g|/g<1-a\,]$ over \mbox{$a\in\{.50,.55,\dots,.95\}$}.
An answer off by half or more scores zero. Place quality is F1 over same-place frame pairs. Both are in per cent; a difference of two scores is absolute; intervals are 95\% bootstrap intervals over streams.

\begin{table*}[t]
\centering\footnotesize
\setlength{\tabcolsep}{4pt}\renewcommand{\arraystretch}{0.9}
\begin{tabular*}{\textwidth}{@{\extracolsep{\fill}}lccc@{}}
\toprule
 & \multicolumn{2}{c}{ARKit-Revisit} & 3RScan-Revisit \\
\cmidrule(lr){2-3}\cmidrule(l){4-4}
Method & test & held-out & test \\
\midrule
\multicolumn{4}{@{}l}{\textit{Blind baselines$^{\dagger}$}~\cite{goyal2017vqav2,chen2024mmstar}}\\
\quad Fitted on training streams & 41.6 [37.7, 45.4]
 & 39.2 [35.4, 43.3]
 & 37.1 [34.1, 40.1]
 \\
\quad Rate $\times$ \method's places$^{*}$ & 39.8 [35.9, 43.4]
 & 36.8 [32.9, 40.9]
 & 32.0 [28.4, 35.8]
 \\
\midrule
\multicolumn{4}{@{}l}{\textit{Adding up parts}~\cite{qharabagh2026lvlmcount,yang2025cambrian}}\\
\quad Sum over visits (privileged) & 26.9 [23.5, 30.5]
 & 30.8 [27.4, 34.3]
 & 17.9 [14.6, 21.3]
 \\
\quad \method without return links & 29.6 [25.8, 33.5]
 & -- & -- \\
\quad \method without return links $\times$ 0.69 & 57.7 [53.9, 61.5]
 & -- & -- \\
\quad Fixed correction, sum over visits $\times$ 0.69 (privileged) & 58.6 [54.8, 62.7]
 & -- & 69.1 [64.8, 73.3]
 \\
\midrule
\multicolumn{4}{@{}l}{\textit{Whole-video reading}}\\
\quad Qwen3.5-9B~\cite{qwen2026qwen35} (same reader) & 51.2 [46.9, 55.8]
 & 44.3 [40.0, 48.9]
 & \textbf{55.1} [49.9, 60.1]
 \\
\quad Gemma~4 12B~\cite{gemma2026gemma4} (other reader) & 27.8 [23.8, 32.1]
 & -- & -- \\
\midrule
\multicolumn{4}{@{}l}{\textit{Other video models}}\\
\quad Cambrian-S-7B~\cite{yang2025cambrian}, whole-video reading & 34.1 [30.3, 38.2]
 & -- & -- \\
\quad Cambrian-S-7B~\cite{yang2025cambrian}, surprise memory & 2.1 [0.7, 3.9]
 & -- & -- \\
\quad VideoAgent~\cite{fan2024videoagent} (amended parser) & 4.9 [3.3, 6.6]
 & -- & -- \\
\quad CountVid~\cite{amini2025countvid} / DirectMe~\cite{wang2026directme}, 40 questions$^{\ddagger}$ & 0.0 / 0.0 & -- & -- \\
\midrule
\method{} (ours) & \textbf{62.5} [58.4, 66.5]
 & \textbf{61.2} [57.7, 65.0]
 & 47.2 [41.7, 52.9]
 \\
\quad with Gemma~4 12B as reader & 51.6 [47.2, 55.6]
 & -- & -- \\
\midrule
\multicolumn{4}{@{}l}{\textit{Oracle (privileged; not ranked)}}\\
\quad \method on true visits & 65.1 [61.6, 68.6]
 & 62.8 [59.2, 66.6]
 & 65.4 [60.9, 70.0]
 \\
\bottomrule
\end{tabular*}
\par\smallskip{\footnotesize --: not run, for lack of time; the pre-registered held-out test has no other video models. $^{\dagger}$\,Highest of eight blind baselines in each column. $^{*}$\,Category rate per place times \method{}'s places. $^{\ddagger}$\,\method: 57.5 [46.7, 68.5] on these 40; timeouts count as zero.\par}
\caption{\method leads on real ARKit returns and falls behind whole-video reading under 3RScan's camera change. Strict MRA (\%) with 95\% bootstrap intervals over streams. Privileged rows use true visits; bold: best row without them.}
\label{tab:main}
\end{table*}

\paragraph{Baselines and controls.}
Many multimodal questions can be answered without the input~\cite{goyal2017vqav2,poliak2018hypothesis,chen2024mmstar,tong2024cambrian1}, so every split reports blind baselines. A blind baseline sees no video content that carries the answer: a constant, a per-category prior, and controls that scale a category rate by duration, recordings or places. All are fitted on training streams of the same dataset.
Whole-video reading gives the same model all 1-fps frames of the stream in one pass, capped at 2,048. \emph{Sum over visits} counts each true visit separately and adds the counts. A fixed correction multiplies this sum by one constant, 0.69, fitted on 48 training streams.
An oracle uses the truth: \method given the true visits in place of its fragments.
Cambrian-S-7B~\cite{yang2025cambrian}, CountVid~\cite{amini2025countvid}, VideoAgent~\cite{fan2024videoagent} and DirectMe~\cite{wang2026directme} run with official code (settings and checks in Supp.~C). CountVid runs on a fixed subset of 40 questions, as many as time allowed (cost in Table~\ref{tab:cost}), DirectMe on the same 40, and timeouts count as zero. VideoAgent runs on all 460 test questions with a Qwen3.5-9B controller; we amended its parser after the first test reads and re-read every answer.
StreamMem~\cite{yang2025streammem} has no released code, and the Spatial-TTT~\cite{liu2026spatialttt} checkpoint fails to run on our streams.

\paragraph{Use of each split.}
Training data set every parameter. ARKit-Revisit test, 3RScan-Revisit and VSI-SUPER-Count guided our failure analysis; only the held-out split and MultiScan-Revisit stayed unseen until evaluated once. Post hoc analyses chose nothing.
\section{Experiments}
\label{sec:experiments}
\paragraph{Key findings.}
Against existing methods, a memory model constructed per video at test time is (1)~more accurate than whole-video reading with the same reader, most at large counts (Sec.~\ref{sec:oneread}); (2)~level as rooms return more or less often, where a fixed correction holds at one return rate only, with frames kept from every stay in our streams (Sec.~\ref{sec:correction}); (3)~ahead of the other memories, which over-count here and cost more (Sec.~\ref{sec:others}); (4)~built with one decision per fragment, enough because a strong match is rarely wrong (Sec.~\ref{sec:bestlink}); and (5)~controlled by one number, the frames kept per place, which sets its memory, input and answer time (Sec.~\ref{sec:size}). (6)~Its limit is a change of camera, which makes places harder to find (Sec.~\ref{sec:behind}).

\subsection{Does Counting by Place Help?}
\label{sec:oneread}
\paragraph{\method outperforms whole-video reading with the same reader.}
On the test split, \method scores 62.5\% MRA (Table~\ref{tab:main}), against 51.2\% for whole-video reading with the same model (paired +11.3\% [5.1\%, 17.2\%]) and 41.6\% for the best blind baseline (+20.9\% [15.7\%, 26.3\%]).
On the held-out split, evaluated once after the method was fixed, it scores 61.2\% against 44.3\% (+16.9\% [11.5\%, 22.2\%]) and 39.2\% for the blind baseline chosen in advance (+22.0\% [17.1\%, 27.0\%]); both gains meet criteria we pre-registered (Supp.~F).
With Gemma~4 12B~\cite{gemma2026gemma4} as the reader of both methods, it scores 51.6\% against 27.8\% (+23.7\% [18.0\%, 29.3\%]), so the gain is not tied to one reader.

\paragraph{Whole-video reading gives a few small numbers.}
On the test split, 87\% of whole-video answers are 1, 2 or 3, and their slope, the least-squares rise in the answer per extra true object (1 for a perfect counter), is 0.32 [0.27, 0.39], against 0.72 [0.65, 0.80] for \method (Fig.~\ref{fig:teaser}, right).
Whole-video reading scores 84.6\% at two objects, where its preferred answer is right~\cite{chapus2026countvideobench}, and 10.5\% at nine or more, where \method scores 56.3\% (+45.8\% [37.1\%, 53.9\%]; Supp.~D.1).

\paragraph{With and without the graph.}
To remove the graph and change nothing else, a same-frame control gives the reader exactly the frames that \method kept, with the same prompts, in one call. On training streams this accumulated memory scores 48.7\% against 64.3\% with the graph, at the same memory and the same 100 thousand input tokens (slope 0.27 against 0.77).
Whole-video reading, with every frame and no graph, keeps more (2.9 against 2.3~MB per minute) and scores less.
The gain comes first from counting in parts: 73\% of visits with a category hold one such object and 89\% at most two, where the reader is best.
Without return links every return is counted again (29.6\% on test), and one graph shared across 40 training videos raises false merges from 4.0\% to 11.7\% and comparisons per fragment from 5 to 194.

\paragraph{Longer streams have larger counts.}
On 30- and 60-minute streams, whole-video reading falls to 41.3\% and 26.0\% and \method scores 68.2\% and 65.6\% (+26.8\% [20.0\%, 33.7\%], +39.6\% [29.3\%, 48.9\%]).
Longer streams pass through more rooms and hold more objects (mean true count 3.6, 5.5 and 6.3), but whole-video reading's answer barely changes (3.2 to 3.5 at true counts of 5 to 8): one pass over many frames falls back on the few small answers the model prefers. Its fall comes mostly from larger true counts; at 60 minutes its 2,048-frame cap also drops frames.

\subsection{Can a Fixed Correction Replace the Graph?}
\label{sec:correction}
\paragraph{On the benchmark's own return rate, it nearly can.}
As every ARKit-Revisit question has an object seen on a return, the sum over visits scores only 26.9\%.
The fixed correction, multiplying this sum by 0.69, lifts it to 58.6\%, which is 3.9\% [$-$0.3\%, 8.0\%] below \method.
The constant is about the share of first visits (5.7 rooms per 8.2 visits), which varies little, so at this rate a constant nearly matches counting by place, also at 30 and 60 minutes (Supp.~D.3).

\begin{figure}[t]
\centering
\input{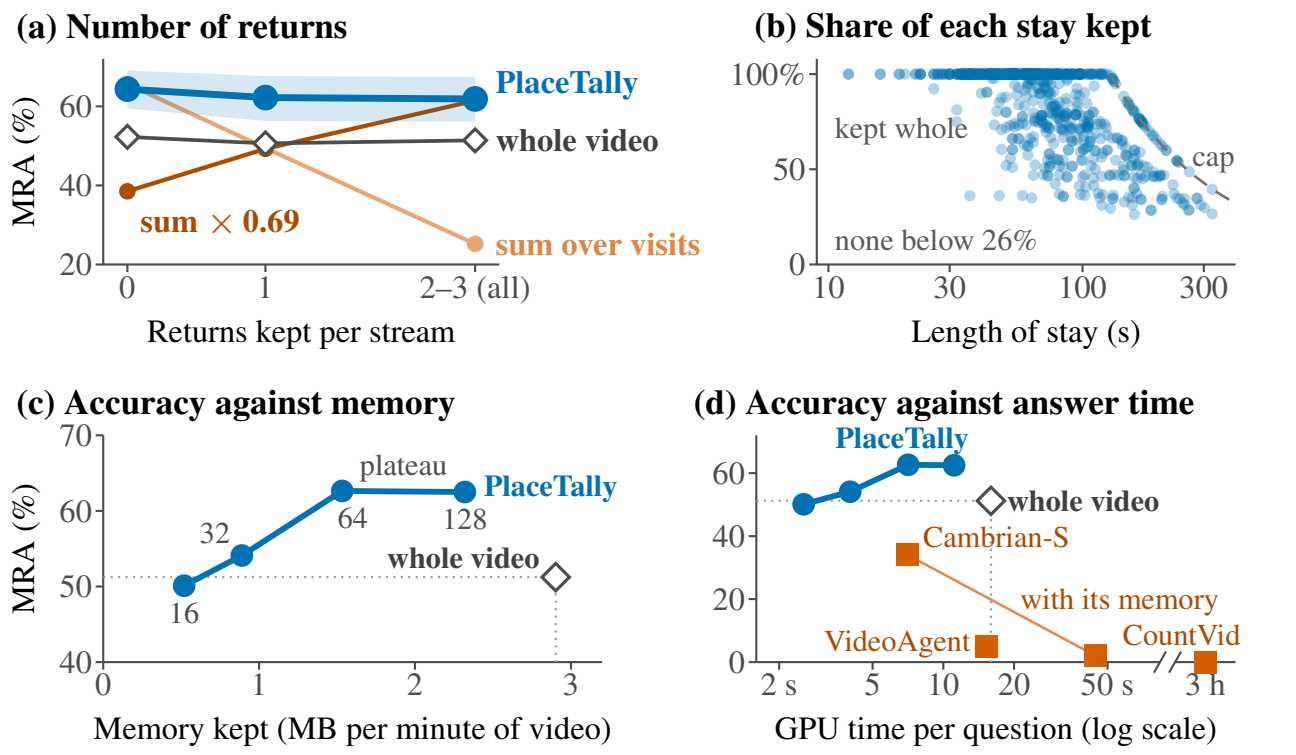}
\caption{Robustness and cost on ARKit test rooms (post hoc, from actual answers). (a) Returns removed from the same 60 test streams; orange: the sum over true visits, as it is and times 0.69 (privileged); band: 95\% interval. (b) The 823 test stays; dashed: the cap of 128 frames per place. (c, d) $K=16$ to 128 frames per place at 11 minutes (gate refitted for 16 and 32); answer time excludes the one-off build (Table~\ref{tab:cost}). Dotted lines pass through whole-video reading: above and left of them is better on both axes.}
\label{fig:curves}
\end{figure}
\begin{table}[t]
\centering\footnotesize
\setlength{\tabcolsep}{2.3pt}
\begin{tabular}{@{}lcrrcrr@{}}
\toprule
 & \multicolumn{3}{c}{ARKit test (598 places)} & \multicolumn{3}{c}{3RScan test (824)} \\
\cmidrule(lr){2-4}\cmidrule(l){5-7}
Variant & F1 & Places & Exact & F1 & Places & Exact \\
\midrule
\multicolumn{7}{@{}l}{\textit{How the stream is cut (same linking)}}\\
Fixed 8 s windows & 32.0 & 4393 & 0 & 59.5 & 1699 & 8 \\
TransNet V2~\cite{soucek2024transnetv2} & 80.7 & 510 & 27 & 48.8 & 423 & 6 \\
SigLIP~\cite{tschannen2025siglip2} similarity drop & 85.7 & 764 & 32 & 70.9 & 2261 & 2 \\
Detector, no continuation & 96.3 & 764 & 51 & 67.8 & 1737 & 3 \\
\midrule
\multicolumn{7}{@{}l}{\textit{How fragments are linked (shared log-odds weights)}}\\
Eq.~\ref{eq:rule} & 96.2 & 595 & 83 & 74.9 & 781 & 39 \\
Solver of Eq.~\ref{eq:partition}~\cite{keuper2015gaec} & 96.6 & 629 & 76 & 76.4 & 1148 & 25 \\
Average link~\cite{jain1999clustering} & 96.5 & 629 & 76 & 75.1 & 1134 & 26 \\
Single link~\cite{jain1999clustering} & 95.1 & 577 & 74 & 71.8 & 621 & 31 \\
\midrule
\method & 96.9 & 594 & 85 & 75.3 & 752 & \textbf{42} \\
\quad with MegaLoc~\cite{berton2025megaloc} returns & \textbf{97.6} & 585 & \textbf{92} & \textbf{76.5} & 738 & 36 \\
\bottomrule
\end{tabular}
\caption{Finding places, post hoc; F1 in \%. Places: predicted (true in the header). Exact: streams with the right number of places, of all 109 and 150. Linking rows share log-odds weights, so Eq.~\ref{eq:rule} there differs from \method, which has its own thresholds. With MegaLoc, only the return evidence changes, its threshold fitted on training streams.}
\label{tab:places}
\end{table}

\paragraph{It holds only at that rate.}
A camera in use need not return at this rate, so we removed returns from 60 random test streams, keeping none, one, or all of them per stream, rebuilt each place graph and counted with the method unchanged (291 questions, Fig.~\ref{fig:curves}a).
Each return shows a room again, often from another side, yet from no returns to all of them \method stays level (64.4\%, 62.2\%, 61.9\%; a change of $-$2.5\% [$-$7.1\%, 2.2\%] on the same questions), while the fixed correction rises by 23.0\% [15.9\%, 29.8\%] (38.5\%, 49.2\%, 61.4\%).
So the correction ties only at the rate it was fitted on, 0.4\% [$-$4.6\%, 5.6\%] below \method on these streams, against 13.0\% [6.7\%, 18.7\%] and 25.9\% [19.1\%, 32.4\%] at one return and none.
Whole-video reading is level but lower (52.3\%, 50.6\%, 51.4\%), and \method's lead over it holds at every rate (10.5\% to 12.1\%, intervals above zero).
A constant fitted on training streams at a mix of these rates (0.81) is still 9.2\% [5.6\%, 12.9\%] below \method, and a per-video rate, \method's places over the visits, needs the place graph anyway (Supp.~D.2).
These results indicate that one number cannot replace a test-time memory model for each video: what can be added up depends on which rooms the camera returned to, and the question-blind place graph works that out before any question is asked.

\paragraph{Short stays are kept.}
Each place's frames are thinned evenly, so on test streams every stay keeps at least 12 frames and 95\% of the shortest tenth are kept whole (Fig.~\ref{fig:curves}b). Questions whose shortest stay with an asked object lasts 42~s or less score as well as those above 67~s (67.8\% and 66.7\%), also at equal true count (Supp.~D.12).

\subsection{Do Other Video Models Close the Gap?}
\label{sec:others}
No: on these streams, the other memories give mostly too-high answers.
VideoAgent scores 4.9\% on all 460 questions, and 304 of its answers are too high.
CountVid scores zero on its 40 questions, against 57.5\% [46.7\%, 68.5\%] for \method: all 36 completed answers are too high, at a median of 16 times the truth, also inside single true places (Supp.~C); the cause remains open.
DirectMe also scores zero: its map holds far more entries than objects before any return, so 38 of its 39 answers are too high.
Cambrian-S reaches 34.1\% reading the whole video and 2.1\% with its surprise memory (Table~\ref{tab:main}). These memories also cost more: VideoAgent keeps 6.7 times the memory and takes 17 times as long to build it, CountVid takes about 3.1 GPU-hours per question, and Cambrian-S, building its memory in each query, 44.4 GPU-seconds against 11.1 plus our one-off build (Fig.~\ref{fig:curves}d).

\subsection{Is One Best Link Enough?}
\label{sec:bestlink}
\paragraph{Cuts and false cuts.}
In Table~\ref{tab:places}, fixed windows and a shot detector miss boundaries that matter, and our detector reaches a place F1 of 96.3\% on ARKit test. The continuation model then gives 85 of 109 ARKit streams the right place count, against 51 without it, with a small gain in counting (+2.0\% [$-$0.6\%, 4.6\%]).

\paragraph{Why the best-link rule is good enough.}
A solver of Eq.~\ref{eq:partition}~\cite{keuper2015gaec} finds no better places than \method (F1 $-$0.3\% [$-$1.9\%, 1.2\%] on ARKit and +1.1\% [$-$3.4\%, 5.2\%] on 3RScan) and it over-splits, with 1,148 places against 824 true on 3RScan (Supp.~D.7).
The reason is that the evidence is one-sided. On ARKit test and 3RScan test, 96\% and 91\% of pairs above $\tau$ show the same place, while 42\% and 67\% of true-return pairs fall below it. A returning fragment's best match in the earlier recording of its room clears $\tau$ in 85\% and 71\% of cases, the mean match in only 74\% and 36\%, so one strong link is the reliable test, and a mean over the place, as in a zone score~\cite{nagarajan2020egotopo}, misses more returns.
\paragraph{Headroom.}
Given true visits, \method rises from 62.5\% to 65.1\% on test and from 61.2\% to 62.8\% on the held-out split, and only 1.9\% [0.7\%, 3.7\%] of objects appear in two predicted places, so the parts are nearly object-disjoint. On training streams, most remaining error is the reader's: with true places it scores 66.5\%, against 89.3\% with exact counts in every gated place (Supp.~D.9).

\subsection{\texorpdfstring{How Much Does the Memory Cost?}{How Much Does the Memory Cost?}}
\label{sec:size}
\begin{table}[t]
\centering\footnotesize
\setlength{\tabcolsep}{2.2pt}
\begin{tabular}{@{}lrrrrr@{}}
\toprule
Method & \shortstack{Build\\speed} & \shortstack{Memory\\MB/min} & \shortstack{Input\\k tokens} & \shortstack{Query\\GPU-s} & \shortstack{MRA\\(\%)} \\
\midrule
\method{}, $K=16$ & 15 & 0.52 & 19.0 & 2.5 & 50.1 \\
\method{}, $K=64$ & 15 & 1.53 & 68.3 & 7.1 & 62.6 \\
\method{}, $K=128$ & 15 & 2.32 & 105.8 & 11.1 & 62.5 \\
Whole-video reading & -- & 2.90 & 69.1 & 15.9 & 51.2 \\
VideoAgent~\cite{fan2024videoagent} & 0.88 & 15.5$^{*}$ & n/c & 15.3 & 4.9 \\
CountVid~\cite{amini2025countvid} & -- & 7.6$^{*}$ & -- & 11{,}084 & 0.0$^{\dagger}$ \\
DirectMe~\cite{wang2026directme} & 0.24 & 3.9$^{*}$ & -- & -- & 0.0$^{\dagger}$ \\
Cambrian-S whole video~\cite{yang2025cambrian} & -- & n/m & n/c & 7.0 & 34.1 \\
Cambrian-S memory~\cite{yang2025cambrian} & -- & 61.9$^{\ddagger}$ & n/c & 44.4 & 2.1 \\
\bottomrule
\end{tabular}
\caption{Cost on ARKit test (11-minute videos), post hoc, on one A100-40GB GPU. $K$: frames kept per place ($K=16$: refitted gate); \method and whole-video reading use Qwen3.5-9B. Build:~$\times$~real time, once per video. Memory: kept after construction (\method: 3.0~MB/min while building). Input: all 460 questions; query: the same 50, nothing cached. --: none. n/c: counted differently. n/m: not measured. $^{*}$\,JPEG estimate (as run: 87, 68, 24~MB/min). $^{\ddagger}$\,GPU state while answering. $^{\dagger}$\,40 questions.}
\label{tab:cost}
\end{table}

\paragraph{One number sets the cost.}
Input tokens are linear in stored bytes ($R^2=0.99$) and answer time is linear in the input ($R^2=0.93$), so $K$ sets all three costs (Table~\ref{tab:cost}; post hoc). With $K=128$, memory grows linearly with the video, at 2.3~MB per minute, and the graph is built 15 times faster than real time, at 11, 30 and 60 minutes alike (Supp.~D.11).

\paragraph{Less memory costs little accuracy.}
With $K=64$, \method reads as much as whole-video reading, stores about half as much and answers 2.2 times faster; it scores 62.6\% against 51.2\%. With $K=16$ it keeps 18\% of whole-video reading's memory and nearly matches it (50.1\% against 51.2\%).
More frames stop helping early: at 11 minutes, $K=128$ takes 1.5 times the memory and 1.6 times the answer time of $K=64$ for no gain (Fig.~\ref{fig:curves}c), and on 30- and 60-minute streams it adds 3.6\% and 3.4\%, intervals reaching zero (Supp.~D.6). As 63 to 65\% of places are read in full at $K=128$, a larger $K$ changes only the rest. About half the input is the presence gate, which adds 2.1\% [$-$0.9\%, 5.3\%].

\subsection{\texorpdfstring{When Places Are Hard to Find}{When Places Are Hard to Find}}
\label{sec:behind}
\paragraph{A change of camera.}
Places stay the right unit under a change of camera, but finding them needs a continuation model fitted to that camera. On 3RScan-Revisit, \method on true visits reaches 65.4\%, above whole-video reading at 55.1\%, while with its own fragments it falls behind (47.2\%, paired $-$7.9\% [$-$14.8\%, $-$1.0\%]; Table~\ref{tab:main}).
Place F1 is 75.3\%, against 96.9\% on ARKit, and 31.1\% [27.2\%, 35.3\%] of objects appear in two predicted places.
The continuation model, fitted on ARKit video, rejoins non-adjacent fragments of one recording 72 to 83\% of the time (99\% on ARKit) and merges adjacent fragments of different rooms 22 to 26\% of the time (3\%); one trained across cameras lost accuracy on ARKit, so we kept ours.

\paragraph{Look-alike rooms.}
On MultiScan-Revisit, place F1 is 90.2\% [83.6\%, 96.4\%]: 97.8\% on ordinary streams and 84.4\% with look-alike rooms, which can be merged (Supp.~E).
\section{Conclusion}
\label{sec:conclusion}
We present \method, a question-blind test-time memory model that constructs a place graph for each video before any question is asked. To count, a memory does not need to know what it is looking for, only where it has been. Experiments on real ARKitScenes returns show that \method outperforms whole-video reading with the same reader on test and held-out rooms, and stays level as the return rate changes, while no single correction holds across rates. One number, the frames kept per place, sets its memory, input and answer time, and it is built 15 times faster than real time.

\paragraph{Limitations and future work.}
Places stand in for object identity while objects stay in their rooms, as our 14 categories do. Our streams join real recordings at a hard cut, and the other return rates come from removing visits from tested rooms; new recordings of continuous walks would test both. The next steps are a continuation model fitted to each camera and return evidence trained on look-alike rooms.

\section*{LLM Usage Statement}
We used generative AI tools (Claude Code and Codex) for code implementation, debugging, for searching related work, and occasionally for suggestions on experimental variations. We also used them to improve writing clarity and grammar, refine figures, and check the manuscript for unclear presentation or misunderstandings. The authors decided the main hypotheses and the research methodology. All AI-assisted work was reviewed by the authors, and we take responsibility for the final content of the paper.

{\small
\bibliographystyle{ieeenat_fullname}
\bibliography{refs}
}

\end{document}